\documentclass{article}
\usepackage{graphicx}
\usepackage[final]{ap_2021}% use [final] for camera-ready
\usepackage[utf8]{inputenc} % allow utf-8 input
\usepackage[T1]{fontenc}    % use 8-bit T1 fonts
\usepackage{hyperref}       % hyperlinks
\usepackage{url}            % simple URL typesetting
\usepackage{booktabs}       % professional-quality tables
\usepackage{amsfonts}       % blackboard math symbols
\usepackage{nicefrac}       % compact symbols for 1/2, etc.
\usepackage{microtype}      % microtypography
\usepackage{todonotes}

\graphicspath{ {./images/} } %image path

\title{Alexa, Let's Work Together: Introducing the First Alexa Prize TaskBot Challenge on Conversational Task Assistance}

\author{
	Anna Gottardi
	\And Osman Ipek
	\And Giuseppe Castellucci
	\And Shui Hu
	\And Lavina Vaz
	\And Yao Lu
	\And Anju Khatri
	\And Anjali Chadha
	\And Desheng Zhang
	\And Sattvik Sahai
	\And Prerna Dwivedi
	\And Hangjie Shi
	\And Lucy Hu
	\And Andy Huang
	\And Luke Dai
	\And Bofei Yang
	\And Varun Somani
	\And Pankaj Rajan
	\And Ron Rezac
	\And Michael Johnston
	\And Savanna Stiff
	\And Leslie Ball
	\AND David Carmel
	\And Yang Liu
	\And Dilek Hakkani-Tur
	\And Oleg Rokhlenko 
	\AND Kate Bland
	\And Eugene Agichtein
	\And Reza Ghanadan
	\And Yoelle Maarek
}

\begin{document}

\maketitle

\begin{abstract}
  Since its inception in 2016, the Alexa Prize program has enabled hundreds of university students to explore and compete to develop conversational agents through the SocialBot Grand Challenge. The goal of the challenge is to build agents capable of conversing coherently and engagingly with humans on popular topics for 20 minutes, while achieving an average rating of at least 4.0/5.0. However, as conversational agents attempt to assist users with increasingly complex tasks, new conversational AI techniques and evaluation platforms are needed. The Alexa Prize TaskBot challenge, established in 2021, builds on the success of the SocialBot challenge by introducing the requirements of interactively assisting humans with real-world Cooking and Do-It-Yourself tasks, while making use of both voice and visual modalities. This challenge requires the TaskBots to identify and understand the user's need, identify and integrate task and domain knowledge into the interaction, and develop new ways of engaging the user without distracting them from the task at hand, among other challenges. This paper provides an overview of the TaskBot challenge, describes the infrastructure support provided to the teams with the CoBot Toolkit, and summarizes the approaches the participating teams took to overcome the research challenges. Finally, it analyzes the performance of the competing TaskBots during the first year of the competition. 
\end{abstract}

\section{Introduction}

%problem and motivation
In the last several years, Conversational Agents such as Amazon Alexa, Apple’s Siri, and Google Assistant have become a popular way for humans to access information. As humans expect to interact with such agents by voice, the conversational assistants must be able to support a broader range of increasingly complex tasks. The Alexa Prize\footnote{\url{https://www.amazon.science/alexa-prize}} is an Amazon Alexa sponsored program that in recent years has enabled hundreds of university students to compete in advancing the state-of-the-art in conversational AI. Since 2016 the SocialBot Grand Challenge hosts a competition among universities across the world to compete in creating the best \textit{SocialBot}, i.e., an Alexa skill that can engage and converse with humans on popular topics and news events for at least 20 minutes. 

However, humans are not only interested in chit-chat. In many cases, humans use the web to find information about activities they can do, for example how to cook a specific dish or tutorials on how to complete a Do It Yourself (DIY) project. Building on the success of the series of Alexa Prize SocialBot challenges, we introduce a novel conversational challenge to develop a TaskBot, i.e., a conversational agent that can interactively assist users in completing real-world tasks. Just as in the SocialBot challenge, the TaskBots must maintain helpful and engaging interactions with the user. However, the purpose of a TaskBot is not merely to converse but to actively assist the user in formulating and accomplishing their task. We can define a TaskBot as a \textit{Conversational Task Assistant} (CTA). Notice that this is quite a different setting from the existing task-oriented assistants in literature \cite{zhang2020,zamani2022conversational}, where the user typically provides information to the task assistant to let the machine perform a task (e.g., booking a hotel, buying flight tickets). A CTA, instead, is meant to provide the humans with both accurate information and assistance about a task that is executed by them.

In the first year of the TaskBot challenge, the participants built conversational agents on top of the multimodal Alexa service with respect to two domains: Cooking and Home Improvement (or DIY). These present different challenges for a CTA that require scientific advancements for a conversational system. For example, the CTA should help the user to find the right task according to their needs; provide information about specific aspects of the task (e.g., ingredients or tools/materials); answer questions the user may ask about the task itself (e.g., how to use a specific tool); identify, structure, and integrate domain knowledge into the interaction; combine task status and dialog state tracking; and generate responses for both voice and visual modalities. These challenges required the participants to invent novel technologies in different areas, like Information Retrieval, Question Answering, Dialog Management, and multimodal interaction. Given that this was the first year of such a challenge, the participants tackled an additional difficulty, i.e., the lack of data. In fact, given the unique setting of a CTA, there is no existing dataset that can provide both information on how users would interact with such a TaskBot and training material.

%approach: TaskBot framework to help university teams innovate
To address these challenges, we released an updated version of the CoBot Toolkit for developing conversational AI agents, which was used previously for the past SocialBot Grand Challenges. The new version of the toolkit was significantly extended for the TaskBot Challenge as detailed in Section~\ref{sec:capabilities}. Using the CoBot Toolkit, the participants were able to spend less time on engineering and focus more on scientific advancements, as overviewed in Section~\ref{sec:scientific_partipants}. 

%user experience and competition phases
TaskBot was launched to a cohort of Amazon employees on September 20, 2021, followed by a public launch on November 1, 2021 at which time all US Alexa users could interact with the participating TaskBots. Upon making a request for task assistance, e.g. \textit{Alexa, Let's Work Together}, Alexa users were connected to one of the ten participating TaskBots. After exiting the interaction with the TaskBot, the user was prompted for a verbal rating: “How helpful was this TaskBot in assisting you?” followed by a task completion prompt: “Were you able to complete your task?” and an option to provide additional free-form feedback. After an initial feedback phase, we held Quarterfinals from December 1, 2021	through January 28, 2022. Nine teams advanced from the Quarterfinals and participated in the Semifinals from February 7, 2022 through March 25, 2022. Five teams qualified to participate as Finalists and participated in an additional feedback phase through May 5, 2022. The closed door Finals were held on May 18-19, 2022.

%contributions this year
The first year of the competition saw rapid progress on the capabilities of the TaskBots, as illustrated by the participants' technical reports as well as the significant improvements in user satisfaction and engagement with the TaskBots as the competition progressed. The Finalists improved their user satisfaction ratings by $19.6\%$ (from 2.7 to 3.23 out of 5) over 24 weeks of the competition. We review the performance of the TaskBots in Section~\ref{sec:results} and discuss insights gathered from the TaskBot Challenge in Section~\ref{sec:conclusions}. 

\section{Capabilities Provided to Teams}
\label{sec:capabilities}

To support TaskBot development, the university teams were provided with unique access to Amazon Alexa resources, technologies, and personnel. The resources are outlined below.

\subsection{Conversational Bot Toolkit (CoBot)}

\begin{figure}[!ht]
\centering
\includegraphics[width=1\textwidth]{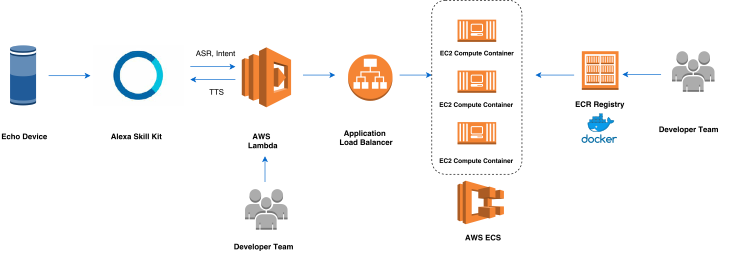}
\caption{CoBot System Diagram and Workflow}
\label{fig:cobot_arch}
\end{figure}

We provided the TaskBot teams with CoBot, a conversational bot toolkit in Python for natural language understanding and dialog management, originally developed for the Alexa Prize SocialBot Grand Challenge \cite{khatri_cobot}. CoBot includes a set of tools, libraries, and base models to help develop and deploy open-domain or multi-domain conversational experiences through the Alexa Skills Kit (ASK, \cite{kumar_ask}) and Amazon AWS (see figure \ref{fig:cobot_arch}). CoBot's modular, extensible, and scalable design provides abstractions that enable the teams to focus more on scientific advances and reduce time invested into infrastructure, hosting, and scaling.

\begin{figure}[!ht]
\centering
\includegraphics[width=1\textwidth]{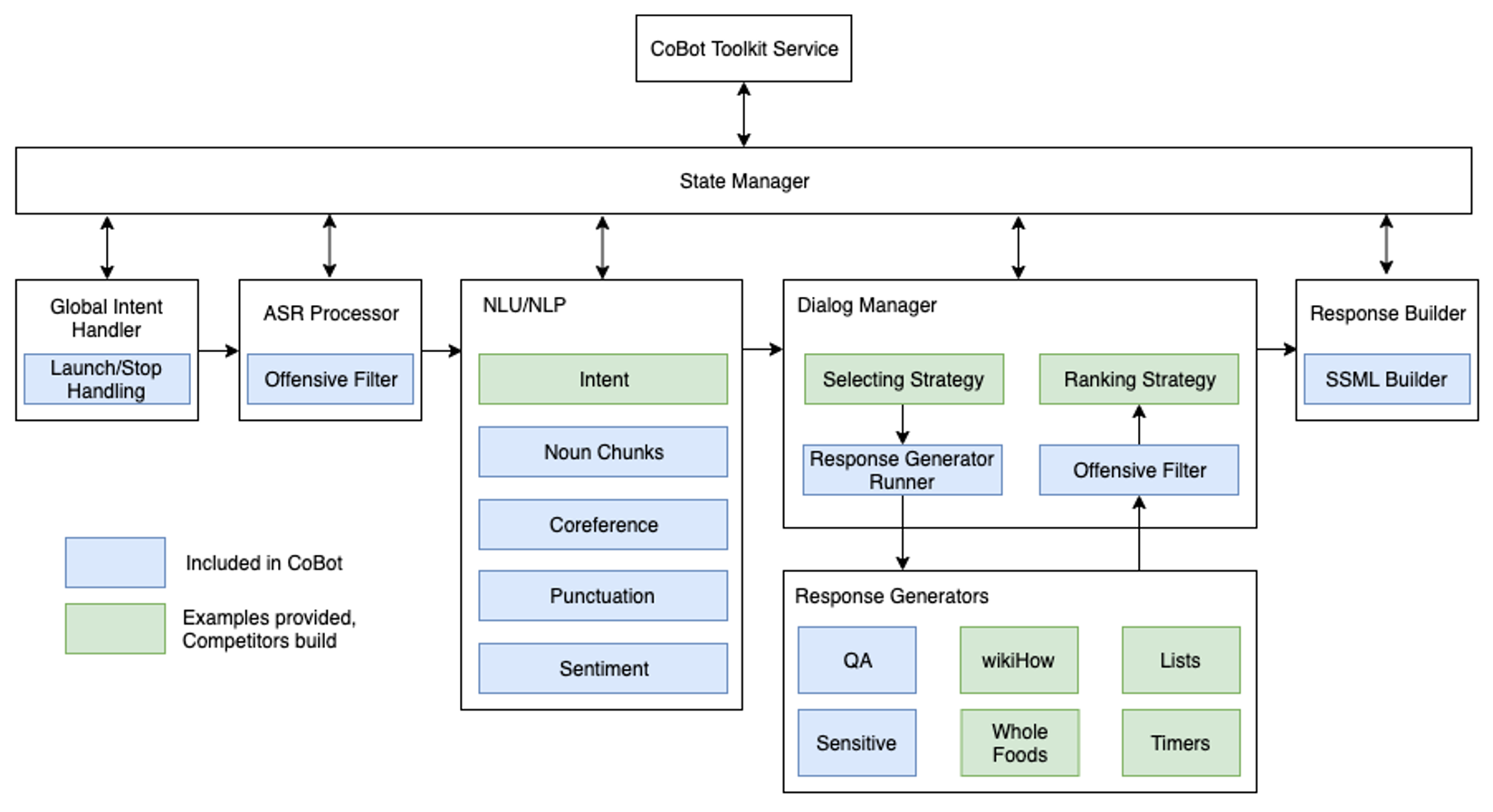}
\caption{Updated CoBot Architecture for TaskBot, new components include wikiHow, Whole Foods, Lists, and Timers}
\label{fig:cobot_detail}
\end{figure}

For the TaskBot challenge, we released a significantly extended version of CoBot with support for long-running tasks and multimodal interactions, integration with ASK APIs for managing shopping lists and timers, and APIs for retrieving Whole Foods Market recipes and wikiHow articles, as illustrated in figure \ref{fig:cobot_detail}. 

\subsubsection{Task Resumption and Completion}

One aspect of TaskBot that differs from the Alexa Prize SocialBot Grand Challenge is the concept of completing real world tasks during the dialog. TaskBot is intended to assist users as they prepare recipes or complete projects around the house. A user who actually performs these activities will require more time between dialog turns and may even need to pause and restart their interaction.

To give users additional time between turns on devices with a screen, we modified CoBot to allow teams to choose when to prompt the user for a response versus when to close the microphone. Closing the microphone removes the pressure on the user to reply immediately and gives them time to look over content shown on the screen. When the user is ready, they use the wake word (such as \textit{Alexa}) to continue their interaction. By default in CoBot, we configured the screen to stay active for 60 seconds before ending the session. Teams have the ability to increase this period up to 30 minutes, to give users more time to gather supplies or perform the current step of the task.

Some tasks require longer breaks between turns, for example when a trip to the grocery or hardware store is needed. To address this case, we allow users to pause their current task and resume the interaction at a later time. When a user ends their interaction by saying \textit{stop}, after collecting their rating we ask, “Were you able to complete your task?” Upon their next invocation of TaskBot, users with tasks marked incomplete will be asked, “Do you want to continue the last task you worked on?” Answering \textit{yes} will allow the user to resume their ongoing task where they left off. An answer of \textit{no} will pair the user with a new TaskBot at random to ensure variety of experience. After 7 days of inactivity, incomplete tasks will expire.

Outside of the skill session, we provide two ways for users to access details about their task. First, we utilize an Alexa feature known as \textit{cards} to provide a summary of the task instructions that can be accessed later on in the Amazon Alexa App, the companion app available for Fire OS, Android, iOS, and desktop web browsers. Second, we also enabled support for the ASK List API in CoBot to allow users to add items to their shopping and to-do lists, which can be accessed in the Amazon Alexa App and across other Alexa-enabled devices. 

\subsubsection{Multimodal Experience}

TaskBot is the first conversational AI challenge to incorporate multimodal (voice and vision) user experiences. In addition to receiving verbal instructions, users with Echo screen devices, such as the Echo Show, could also be presented with step-by-step instructions, images, or videos that enhance task guidance.

To enable developers to build interactive voice and visual experiences, the Alexa Skills Kit provides a visual design framework called Alexa Presentation Language (APL). APL includes visual elements that scale across device types and can support both voice and touch interactions. For the TaskBot competition, we provided teams with three APL responsive templates: Alexa Text List, Alexa Image List, and Alexa Detail. A responsive template is a complete viewport design that includes the background, header, and content. For example, the Text List template presents a scrolling list of text items with a background and header. Responsive templates simplify the developer experience by providing built-in components and automatic support for different screen sizes.

\begin{figure}[htb]
\centering
\includegraphics[width=1\textwidth]{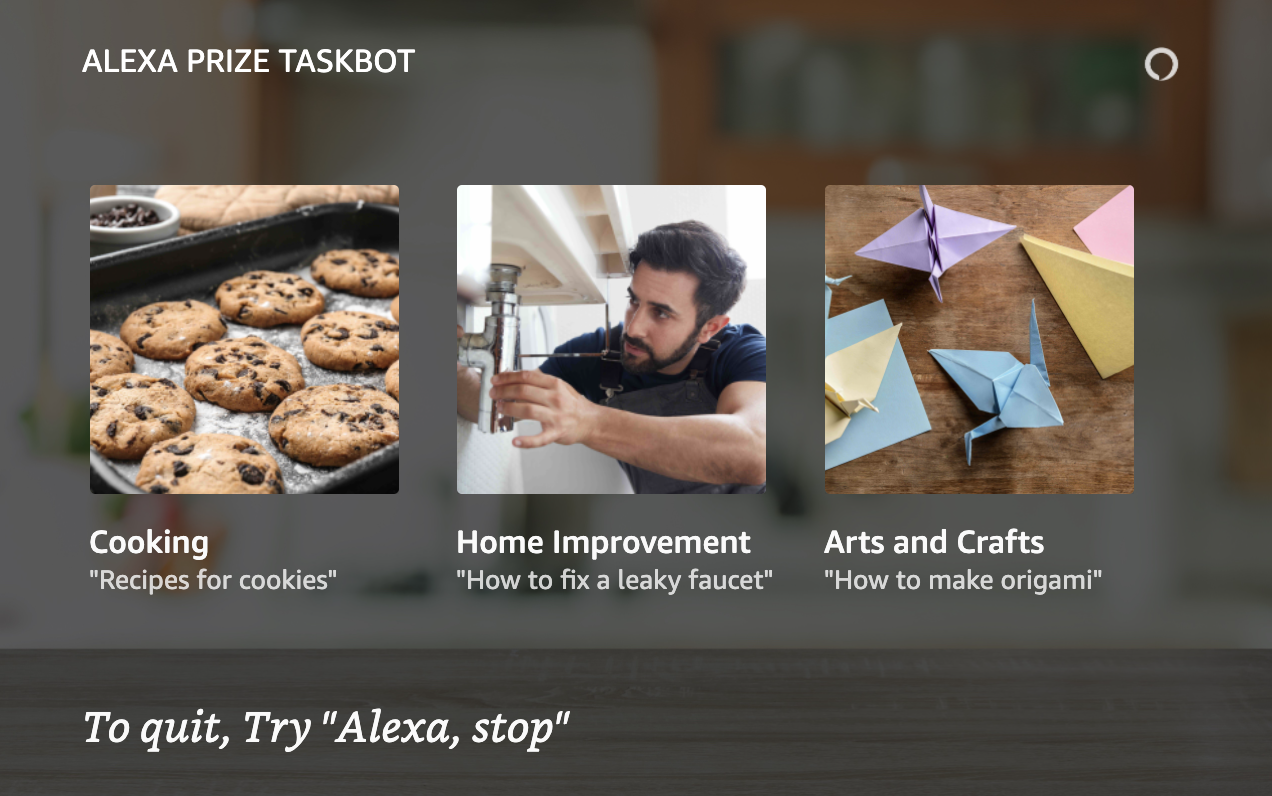}
\caption{An Alexa Prize TaskBot welcome screen, illustrating the TaskBot APL Template provided to the participating teams}
\label{fig:cobot_apl}
\end{figure}

We restricted teams to use the provided templates in order to provide a uniform experience for users and reduce the effort teams needed to spend on building new layouts from scratch. This restriction meant the Alexa Prize team needed to react quickly to evolving needs by making additional templates available throughout the competition. We introduced a video template to support playing how-to videos and recipe walkthroughs via a clickable media player. We also added a welcome page template based on our observation that users benefited from having a landing page that introduced the bot's capabilities.

While the multimodal experience was a major focus of the TaskBot competition, many Alexa users also engaged with TaskBot on devices without screens (known as headless devices). Teams received a flag in their skill requests to indicate whether the user's device supported APL and tailored the voice experience accordingly.

\subsection{Automatic Speech Recognition and Text to Speech}
We provided Automatic Speech Recognition (ASR) to convert user utterances to text and Text-To-Speech (TTS) to render text responses from TaskBots to users via voice. Our ASR model is tuned for conversational data and features custom end-pointing and extended recognition timeouts for longer free-form interactions. Alexa Prize teams also received access to tokenized n-best ASR hypotheses, including confidence scores for each token. 

For the TaskBot challenge, we extended the Alexa Prize Conversational ASR Language Model to incorporate TaskBot utterances transcribed over the course of the competition, as well as new datasets for recipes and wikiHow article titles. On a TaskBot-specific test set, we achieved a $27.8\%$ reduction in word error rate relative to our existing model. We plan to deploy the improved ASR model for all Alexa Prize interactions at the end of the first TaskBot challenge.

\subsection{Infrastructure}
We provided free Amazon Web Services (AWS) to teams, including but not limited to: GPU-based virtual machines for building models, SQL/NoSQL databases, and object-based storage with Amazon S3. We also provided load testing and scalability tools and architectural guidance.

\subsection{Customer Feedback Data and Evaluation Metrics}
Both subjective satisfaction ratings and task-oriented metrics of task completion and resumption (described in section \ref{sec:results}) were collected. The TaskBot teams received these metrics for their own bot and an anonymized leaderboard with average metrics for all bots. In addition, teams received transcriptions of freeform feedback provided by users at the end of interactions with the team’s bot, which allowed the teams to gain qualitative insights into the users' impression of the TaskBots.

\subsection{Support}
In addition to data and infrastructure, we engaged with university teams in several ways to provide support and feedback:
\begin{itemize}
    \item Hands-on bootcamp training material and best practices design guidelines.
    \item An Amazon-only internal beta phase, to provide traffic from Amazon employees to help inform and improve bot performance before general availability to all Alexa users.
    \item Detailed report cards on bot experiences prior to public launch, evaluating functionality as well as the bot's ability to maintain anonymity and handle inappropriate interactions.
    \item Weekly office hours for 1:1 consultations with a dedicated Alexa Prize Solutions Architect, Program Managers, UX Designers, and members of Alexa science and engineering teams.
    \item A virtual summit bringing together university teams with Alexa scientists, engineers, and senior leadership for keynotes and technical deep dives. 
    \item On-demand access to Alexa Prize personnel via Slack and email.
\end{itemize}

\section{Scientific Advancements}
\label{sec:scientific_partipants}

During the challenge, the participants had to develop their bots in order to satisfy user needs in performing Cooking or Home Improvement tasks. To achieve an acceptable level in the user experience, the participants developed specific modules to address Natural Language Understanding (NLU), Information Retrieval (IR), Ranking and Recommendation, Response Generation, and Dialog Management. An illustrative example of a voice-based interaction between one of the finalist TaskBots and one of the Alexa Prize team members is shown in Figure~\ref{fig:example-conversation}. This interaction demonstrates the complexity of the overall task, and some of the design decisions the participating teams have taken to facilitate the users' interest and engagement, while progressing on the desired task.

\begin{figure}[htb]
\scriptsize
\center
\begin{tabular}{|p{0.2cm}|p{0.9\linewidth}|}
\hline
A & Hi, this is an Alexa Prize TaskBot. I'm excited to help you cook delicious recipes, and handy DIY tasks!  Try asking me things like: How to bake a cake, or, How to Decorate a Flower Pot. Or better yet! Try our spring suggestions!\\
\hline
U & I want to cook\\
A & I love cooking! If you want anything in particular ask me for it. You can also try our suggestions!\\
\hline
U & What are your suggestions?\\
A & Here is our special mix of Spring suggestions, courtesy of Whole Foods Market. No-Bake Spring Berry Pie, Avocado Salmon Kale Salad, and Rolled Ice Cream.  The  third  recipe is very popular,  hundreds of  people have tried making it.  You can also ask me for more options.\\
\hline
U & Select rolled ice cream\\
A &  This is making my mouth water.  The recipe Rolled Ice Cream has a 4.2 star rating. and  Serves 4.  To begin with, I suggest you prep the ingredients. If you want, you can ask me for the `ingredient list', otherwise, ask me to begin this delicious recipe by saying `start recipe'.\\
\hline
U & Tell me the ingredients\\
A &  Rolled Ice Cream is just like momma's cooking: simple yet delicious, it uses 4 ingredients. They are on the screen.  If the list is too long, you can scroll down to see them all or ask me to read the ingredients.\\

\hline
U & Read the ingredients\\
A & I will be reading them shortly, feel free to interrupt me at any point.  1 pint heavy cream. 1 can sweetened condensed milk  . Pinch of salt. Topping of choice.  Do you have all of that? You can ask me to repeat or, when you're ready, just say, start recipe.\\

\hline
U& I don't like heavy cream\\
A& You can substitute 1 cup of cream with 0.75 cups of milk and 0.25 cups of margarine. Melt butter and let it cool. Then mix both ingredients together.\\

\hline
U& Start cooking\\
A& Rolled Ice Cream will be  delightful ,  but please be careful when using any tools or equipment. Remember,  safety first!  Step 1.  Whisk the heavy cream, condensed milk, and salt in a bowl.   To keep going say next!\\

\hline
U & Next\\
A & This is getting interesting! Now comes step 2 .  Pour the mixture over a half sheet pan, or multiple quarter sheet pans. Make sure the ice cream base just barely covers the bottom. It should be ¼-inch (6 mm) thick or less. and look, would you like me to astonish you with a fun fact?.\\

\hline
U & Sure\\
A & Curiosity time: In the United States, on the 17th of July, it's the national ice cream day. I found this really interesting! Anyway, to continue, say `next'.\\

\hline
U& Go to the last step\\
A& Step 6.  This is the last step of this recipe.  Place the rolled ice cream in a chilled bowl and top with any toppings you desire.\\
\hline
U& Next\\
A& Excellent work, we just finished Rolled Ice Cream together!  And to end this on a great note, here's a fun fact about it: In the United States, on the 15th of January, it's the national strawberry ice cream day.  Now you have both food for thought and for your starving belly!  Hope you had fun with me today! Say finished, if you're all done for now.\\
\hline
U& Complete\\
A &I really enjoyed working with you today! I hope I can help you again soon! \\
\hline
\end{tabular}
\caption{An example interaction of an Alexa Prize team member (U) with one of the TaskBots (A)}
\label{fig:example-conversation}
\end{figure}

Given that this year's TaskBot challenge was the first of its kind, all the participants had to invent ways of overcoming different problems, such as scarcity of training data for statistical models and a general lack of clarity on how users would interact with a TaskBot. These aspects can significantly affect the system design and the implementation of machine learning based solutions. In this section, we summarize the main scientific advancements the participants explored when implementing their bots. Each participating team described their innovations in more detail as part of these proceedings.

\subsection{Natural Language Understanding}
\label{nlu_teams}
%@Oleg

%ASR Correction, Domain Classification, Intent Classification, Harm Classification, Semantic Parsing,

As the Natural Language Understanding (NLU) is the core of any conversational system, the teams put significant effort into designing and implementing their NLU modules. Most teams chose a standard sequential NLU approach of domain classification, followed by intent classification and semantic parsing. Several teams also implemented ASR hypothesis selection or correction models by either re-ranking provided hypotheses or using rule-based approaches to correct them. For domain/intent classification tasks most teams used pre-trained transformer-based models (e.g. BERT or RoBERTa) fine-tuned on the target domain data. Several teams further combined intent classification with semantic parsing in one task. For example, \textit{QuakerBot} used generative pre-trained transformer model (GPT-3) to generate the output sequence of intent and slots. {\em TacoBot} developed a hierarchical intent recognition model, framed as a multi-label classification task filtered by dialog state. To train the model, extensive data augmentation via synthetic dialog generation was used, as well as leveraging existing state of the art contextual language models prompted with the intent descriptions and the interaction context. {\em GauchoBot} performed intent classification and slot filling together, with a model trained with synthetically generated utterances. The team also introduced an intent entailment task to improve the robustness of the intent prediction.

In addition a few teams decided to extend their NLU systems with additional capabilities. For example, \textit{QuakerBot} used sentiment analysis to evaluate the user feedback when asking about their progress in the task, and \textit{TWIZ} introduced clarifying questions in a non-intrusive manner using gamification.

Finally, all the teams implemented harm classification models, in which architecture varied from a simple regex approach to transformer-based models.

\subsection{Search and Recommendation}
\label{search_teams}
%@Eugene

Helping the user to formulate their task request and identifying the relevant task is a critical step in assisting the user.  For search, a common and effective approach among the teams consisted of two-stage task retrieval and matching: first, a high-recall lexical retrieval using either provided APIs or external search; followed by semantic re-ranking. Both steps were necessary for flexible matching of user requests to the available tasks. 

For the first retrieval stage, most teams used some form of automatic query expansion. For example, {\em TacoBot} used aggressive query expansion to improve search result recall, which could then be refined, while other teams, such as {\em Maruna} also incorporated pseudo-relevance feedback to further broaden the query. {\em Howdy Y'all} used both query expansion and rewriting. 

The second sage of retrieval used more sophisticated semantic re-ranking methods, using variants of BERT models fine-tuned on relevant data. For example, {\em Maruna} used Distilled-BERT, {\em TWIZ} used DPR with a dual encoder, while {\em QuakerBot} used SentenceBERT. These techniques allowed the teams to identify the small set of results to display for the user. 

Since users often might explore and iteratively refine their task needs, both search and proactive recommendation was necessary. For example, {\em Howdy Y'all} used explicit user preference elicitation when the query is too general by using a conversational recommendation system, which either prompted the user for more preferences or recommended a small set of items. Since asking for user's preferences can cause friction, teams came up with creative ideas to make clarification questions more fun. For example, {\em TWIZ} introduced a gamification of the clarifying questions by creating The Recipes game, to challenge a user to think of a recipe, and narrow down the search space as they asked the user clarifying questions until the system could guess which recipe the user had in mind.

\subsection{Knowledge}
\label{knowledge_teams}
All of the teams performed extensive pre-processing of underlying data to represent the tasks as some form of a graph, such as a sequence of steps, or more complex topologies. For example, {\em GRILL} developed \textit{TaskGraphs} - a dependency acyclic graph structure to represent the relationship between the actions a user needs to complete a task. The TaskGraphs were built automatically from the textual task descriptions and task steps, if available. Other teams such as {\em Howdy Y'all} built databases of tasks and recipes, recording attributes like quantity and temperature.  Other teams, such as {\em TWIZ}, decomposed the task steps into smaller, digestible units, which enabled more natural and interactive instructions, and associated task steps with visual illustrations. Many teams, such as {\em GauchoBot}, augmented the standard wikiHow tasks and recipes with external sources. 

One critical use of knowledge is answering user questions.  Many teams combined general question answering (QA) modules with structured QA over knowledge bases, with some teams contextualizing the QA task by encoding the interaction and task information. 
%For example, {\em GauchoBot} proposed four QA modules, one each for QA over structured knowledge content, one for contextualized QA, and one for unstructured domain knowledge, augmented with general domain QA module. 
%@Eugene
%Task Representation, Task knowledge base construction, 
%GRILL: representation for task structure called TaskGraph. This is a dependency acyclic graph representing a model of the actions the user takes to complete a task. The Taskgraphs are built starting from crawled websites contents. 

\subsection{Response Generation}
\label{response_teams}
%@Giuseppe

When it comes to providing responses to the users, the participants typically relied on the architecture provided by the CoBot framework. A set of responders is selected to produce a response given the latest user utterance, and then a single one is selected among produced responses.
Each responder realizes a specific functionality (e.g., greetings, search, question answering, step-by-step instructions). We observed a plethora of different techniques borrowed from the NLP field in order to implement the responders. For example, the \textit{Condita} team implemented two types of response generation methods: stateful and stateless. The former can change the state of the dialog manager (e.g., Launch, Search, Task Selection, etc.). The latter, instead, cannot change the state of the dialog manager (e.g., open-domain QA, help responder, etc.). 
Many teams adopted neural-based approaches: due to the lack of training examples, the main usage of such models is with few-shot learning along with large pre-trained language models. For example, the \textit{QuakerBot} team adopted the GPT-3 model \cite{brown_gpt3} in different responders to produce utterances for their system, e.g. for question answering. Some teams also used summarization models (e.g., \textit{QuakerBot} with GPT-3) to shorten the length of the responses, especially when dealing with recipe or task instructions for devices without a screen. This is crucial to provide to the user with responses that are not too long and that don't contain too much information that can be hard to grasp. In contrast, the \textit{TWIZ} team introduced “curiosities” or trivia into the interaction to increase the user engagement, retrieving and adding contextually relevant trivia to liven up the responses. 

% QuakerBot: summarization of long steps with GPT-3

% \textbf{Condita:} They have to type of responders - stataful and stateless. Stateful responders change the state in the DM state machine (Launch, Search, Task Selecation, etc.).Stateless responders don't change the state (help responder, open-domain QA, sensitive responder, shopping list responder, timer responder). 

\subsection{Dialog Management}
\label{dm_teams}
%@Giuseppe

In conversational systems the Dialog Manager (DM) is the component that 1) keeps track of the information provided in previous turns of the interaction (i.e., State Tracking) and 2) decides which action should be executed given the state and the current user utterance (Action Prediction). Due to the scarcity of data for the first year of the challenge, most of the teams adopted rule-based approaches to dialog management (e.g., \textit{Maruna}, \textit{GRILL}, \textit{Tartan}, \textit{Condita}). For example, the \textit{GRILL} team represented the interaction in their DM as a finite-state machine, where each state represents a different user goal (e.g., planning, execution, farewell). The state is then used as input to the Neural Decision Parser (cfr. \ref{nlu_teams}) along with the current context to output the actions and argument to be executed.
The \textit{Condita} team implemented a manually defined state-machine for their DM. At each state they can invoke multiple response generators to produce a set of responses. The final response is selected through a neural response reranker model implemented using BERT-FK, which was trained using stack-exchange cooking and DIY related questions and Alexa Prize interactions to further fine tune the model.

Some teams tried to detect special actions to make the interaction more appealing for the user. For example, the \textit{Howdy Y'all} team implemented an empathetic responder which is triggered when the interaction is considered to be going poorly by the DM. In particular, they predict the rating of an in-progress interaction by considering features like the length of the interaction and the number of times a fallback response is provided. If the predicted rating is lower than a threshold, the DM chooses the empathetic responder that will apologize to the user to de-escalate the situation. {\em TWIZ} injected relevant trivia to increase engagement.

Another aspect some of the participants tried to take care of is detecting when to ask clarification questions. For example, \textit{Maruna} detects when to ask clarification questions and in that case they generate closed-domain and open-domain clarification questions. Similarly, \textit{GRILL} detects if a user needs guidance when searching for a recipe or an article and it initiates a sub-dialog to elicit preferences from the user. \textit{TWIZ} incorporated clarification questions as a game, to both clarify intent and increase engagement.

\subsection{Question Answering}
\label{qa_teams}
%@Eugene

Answering user questions (QA) during task exploration and execution was prioritized by all the teams. The CoBot Toolkit provided an API for open-domain QA for general questions that accesses Alexa’s semantic knowledge graph (also used in production to answer Alexa questions) which all the teams used as a fallback. 

For task-specific questions, many teams used variations of pre-trained transformer language models, and fine-tuned on manually annotated TaskBot data. For example, {\em TacoBot} and {\em GRILL} adapted the UnifiedQA model, which then was fine-tuned for the TaskBot context data using manual annotations. {\em Maruna} used a pre-trained RoBERTa model, augmented with a Web-based QA system for passage retrieval and re-ranking.

Other teams developed specialized models for different types of questions. For example, {\em  QuakerBot} handled 5 types of questions (Ingredient, Context Dependent, General, Step Related, Article Related), which were classified with a combination of rules and GPT-3 intent detection. %General questions handles with Evi API. The others with information extraction from the recipe or instructions. If no one of these approaches works, they try to use a GPT-3 model with prompting to extract the answers from the recipe/task document. 
Other teams, such as {\em Condita} experimented with summarization models to shorten the responses, and {\em Howdy Y'all} explored a generation model trained to automatically reconstruct answers to questions.

\subsection{Training and Data Generation}
\label{training_teams}
%@Giuseppe

In order to support advanced natural language interaction, the participants needed to generalize over purely rule-based approaches. To do so, they needed to adopt machine learning techniques. Given that this was the first year of the TaskBot challenge, the teams faced the problem of data scarcity. For this reason, the participants adopted different strategies in order to enable the usage of machine learning models within their bots. 
Many teams manually curated training datasets for specific problems. For example, the \textit{Condita} team manually added fun facts about food in order to provide additional facts about a recipe. Also, the \textit{GRILL} team adopted a human-in-the-loop approach starting from 50 examples to devise 600+ examples to train their Neural Decision Parser.
In general, the participants tried to re-use as much as possible what is available on the web. Different teams (e.g., \textit{Condita}, \textit{GRILL}) built weakly supervised training material from different websites. Also, some teams adopted existing datasets to train NLU models. For example, \textit{Maruna} adopted the MIMICS \cite{MIMICS} dataset to train a facet extraction system and CLINC \cite{CLINC} to train an intent classification model.

The data scarcity problem affected also the choice of the models to be used. Most of the teams opted for using large pre-trained language models as the base for their models. This enabled them to use different techniques for zero-shot or few-shot learning. For example, \textit{QuakerBot} adopted GPT-3 \cite{brown_gpt3} like models with prompting in a few-shot learning setting for both intent detection and slot filling. Also, the \textit{GRILL} team adopted a zero-shot paradigm for their internal QA module using the UnifiedQA model \cite{2020unifiedqa}.

\subsection{Multimodal Experience}
\label{multimodal_teams}
%@Oleg

While all the teams were required to support a multimodal experience using both verbal and visual information, some went an extra step in this direction by incorporating videos in the beginning of the task, balancing information presented on the screen vs information provided via voice, and highlighting important snippets of the text. Several teams introduced an adaptive bot behavior based on the device type. For example, for headless devices \textit{Condita} extended their verbal prompts with information which is normally shown on screen (like recipe rating or time to complete the task). Other teams experimented with visual effects in images and videos. For example, \textit{TWIZ} used a 3D Ken Burns effect, where the video zooms out and pans; as the zoom progresses, the distance between elements increases, thus suggesting a 3D illusion.

\section{TaskBot Performance: Results and Analysis}
\label{sec:results}

Building on the success of the SocialBot challenge, the user's explicit ratings and feedback were used to evaluate the TaskBots. Additionally, task-oriented measures of task completion and task resumption were introduced, allowing a more direct way to evaluate the TaskBots on their effectiveness in assisting the users. Over the course of the competition, the TaskBots demonstrated significant improvements in user experience. In this section, we provide various metrics to evaluate the TaskBots’ performance in the first year of the competition, including comparisons between the Finalists and all TaskBots.

\subsection{Satisfaction Ratings}
\label{sec:ratings}

\begin{figure}[!h]
\centering
\includegraphics[width=1\textwidth]{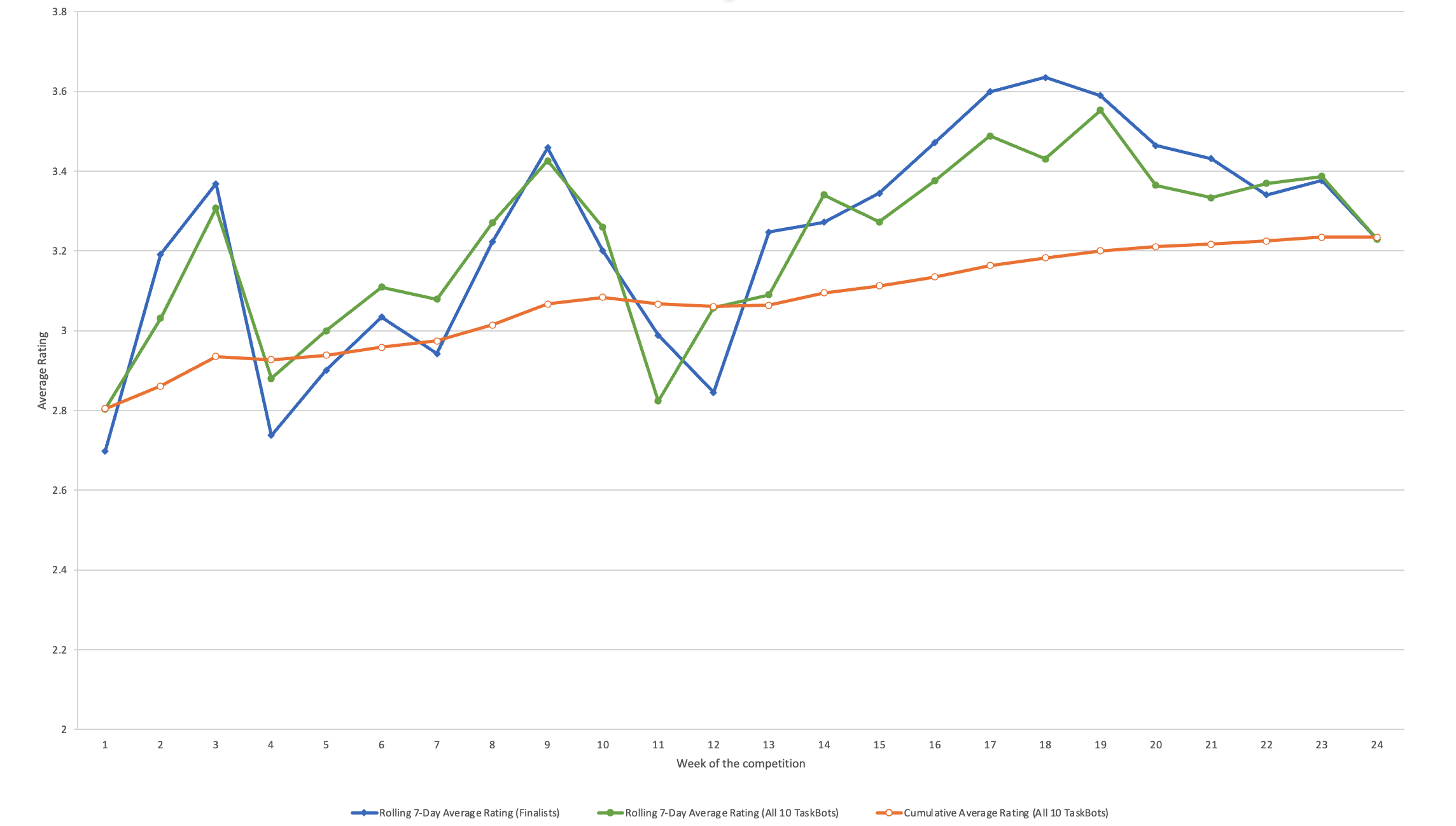}
\caption{Rolling 7-Day Average Rating of User Satisfaction over the period of the competition for all TaskBots (Green), Finalists (Blue) and the progression of the cumulative ratings for all TaskBots (Orange). }
\label{fig:l7d-ratings}
\end{figure}

The primary mechanism of evaluating the TaskBots was explicit user satisfaction ratings. After each interaction, Alexa users were asked to rate their interaction with the TaskBot on a scale of 1-5, according to the prompt, “How helpful was this TaskBot in assisting you?” Please note that the TaskBot rating prompt differs from the prompt used in the SocialBot competition (“How do you feel about speaking with this SocialBot again?”) and thus the ratings should not be compared directly between the two competitions. As shown in Figure~\ref{fig:l7d-ratings}, the Finalists improved their rolling seven-day average ratings by $19.6\%$ (from 2.7 to 3.23) over 24 weeks of the competition. The cumulative average rating across all teams also increased $15\%$ during the course of the competition, from 2.8 to 3.23. 

\subsection{User Satisfaction on Multimodal vs. Headless Devices}
\begin{figure}[!h]
%\caption{Cumulative Multimodal vs Headless Ratings}
\centering
\includegraphics[width=1\textwidth]{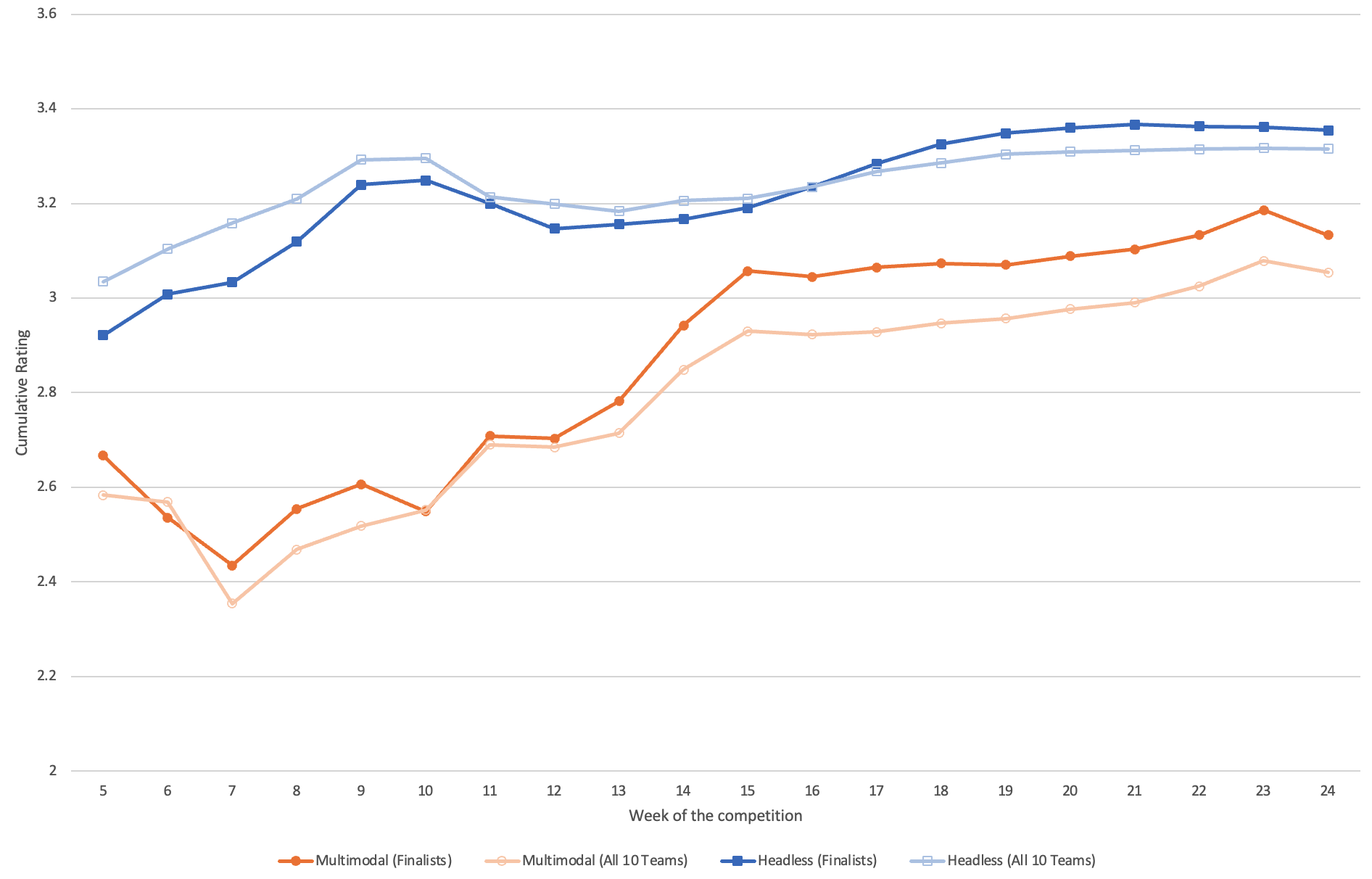}
\caption{Progression of user satisfaction ratings (cumulative) for interactions on devices with a screen (Multimodal) vs. devices without screens (Headless).}
\label{fig:device-ratings}
\end{figure}

The TaskBot challenge is the first Alexa Prize challenge that supports multimodal (voice and vision) user experiences. In Figure~\ref{fig:device-ratings} we show the progression of satisfaction ratings separated by device type, where Multimodal indicates devices with a screen and Headless indicates devices without a screen, beginning Week 4 of the competition. Finalists started with a 2.67 average rating for multimodal devices, and over the course of the competition, they were able to improve their cumulative average by $17\%$, to a final rating of 3.13. 

\subsection{Task Completion and Resumption}
In addition to the satisfaction ratings, task-oriented metrics of task completion and resumption were introduced. At the end of each interaction, once the user provided their rating, they were asked, “Were you able to complete your task?” If the user responded positively, the task would be considered {\em complete}. If a user invoked TaskBot more than once, and had a previous incomplete task  pending, they were asked, “Do you want to continue the last task you worked on?” If the user responded positively, the task would be considered {\em resumed}. 

\begin{figure}[!h]
\centering
\includegraphics[width=1\textwidth]{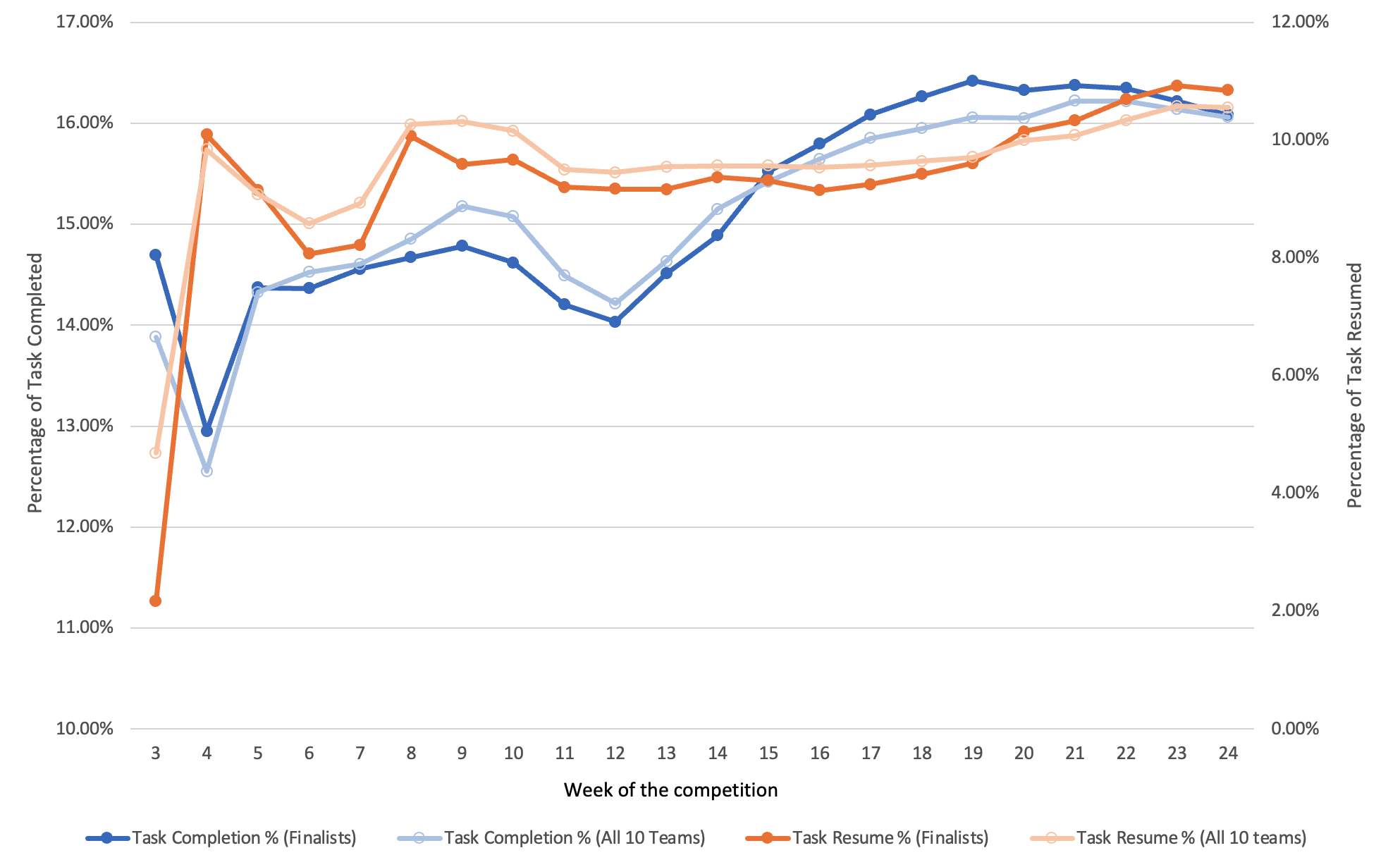}
\caption{Progression of Task Completion (left axis) and Task Resumption (right axis) over the period of the competition.}
\label{fig:completion-ratings}
\end{figure}

The statistics of the resumed and completed tasks were collected starting Week 3 of the challenge and are reported in Figure~\ref{fig:completion-ratings}. The task completion average represents the number of \textit{Yes} responses to the completion question divided by (total number of interactions - number of interactions with \textit{Yes} responses to the resume question), i.e. we do not include resumptions, only the original task. The task resumption average represents the number of interactions with \textit{Yes} responses to the resume question divided by (total number of interactions - number of \textit{Yes} responses to the completion question). In other words, we wish to capture the percentage of incomplete interactions which the user chose to resume.

On average, over the whole competition, $16\%$ of users reported that they were able to complete their task and over $10\%$ of users resumed their previous incomplete task. The completion rates for both Finalists and all teams increased as the competition progressed overall, and consistently after week 12. This indicates that the TaskBots were increasingly able to facilitate the users' tasks.

%Percent Completed and Resumed combined
%\subsection{Percent Resumed}
\subsection{Interactions Analysis}

In this section, we report some analysis on the quality of the interactions that users had with the TaskBots. Firstly, we observed that around $68\%$ of the interactions fell into the Cooking domain, while the remaining $32\%$ fell into the Home Improvement (DIY) domain.

Within a single interaction we can distinguish 2 phases: i) searching, where the user is exploring different tasks; and ii) executing, where the user chose a task and is now actually executing the task. We detected how many turns users spent searching or executing during their interactions. In the first half of table \ref{tab:search_vs_execution} the statistics on search vs. execution for all interactions are reported. Notice that most of the turns are spent in the search phase ($5.47$ turns on average) rather than the execution ($2.39$ turns on average). This trend is less evident for the Home Improvement domain, where the difference between search and execution is lower ($6.13$ and $4.23$ average turns for search and execution, respectively). This could be explained by the fact that the Home Improvement tasks from wikiHow provide additional and interesting information that result in a higher engagement from users. On the contrary, the amount of information available for each step of a recipe was typically less.

\begin{table}[htb]
\centering
\begin{tabular}{|l|cc|cc|}
\hline
 & \multicolumn{2}{c|}{\textbf{All Interactions}} & \multicolumn{2}{c|}{\textbf{Completed Interactions}} \\ \hline
 & \multicolumn{1}{c|}{\textbf{\begin{tabular}[c]{@{}c@{}}Avg Search\\ Turns\end{tabular}}} & \textbf{\begin{tabular}[c]{@{}c@{}}Avg Execution\\ Turns\end{tabular}} & \multicolumn{1}{c|}{\textbf{\begin{tabular}[c]{@{}c@{}}Avg Search\\ Turns\end{tabular}}} & \textbf{\begin{tabular}[c]{@{}c@{}}Avg Execution\\ Turns\end{tabular}} \\ \hline
\textbf{All} & \multicolumn{1}{c|}{5.47} & 2.39 & \multicolumn{1}{c|}{5.83} & 5.50 \\ \hline
\textbf{Cooking} & \multicolumn{1}{c|}{5.17} & 1.53 & \multicolumn{1}{c|}{6.20} & 4.70 \\ \hline
\textbf{Home Improvement} & \multicolumn{1}{c|}{6.13} & 4.23 & \multicolumn{1}{c|}{5.27} & 6.73 \\ \hline
\end{tabular}
\caption{Search vs. Execution turns statistics for all interactions and for those interactions for which the task or recipe was reported as completed.}
\label{tab:search_vs_execution}
\end{table}

To further understand the users' behavior with respect to the search and execution phases, we report the analysis also on those interactions for which the task or recipe was reported as completed by the users. In this case, the domain distribution is a bit different, i.e., about $60\%$ of the interactions are in the Cooking domain. In the second half of Table~\ref{tab:search_vs_execution} we report the statistics for the completed interactions. In this case the average number of turns for the search and execution phases is similar ($5.83$ and $5.50$ for search and execution, respectively). Still, the number of turns for the execution phase is lower than expected if we compare it to the number of steps a typical recipe or task is made of. This could be explained by the fact that users may not really be executing a task or recipe. Instead, they are still exploring how a task or a recipe should be executed and, probably, after a few steps may want to change the task/recipe.

\section{Discussion and Conclusions}
\label{sec:conclusions}

% First paragraph
As this was the first year of the TaskBot challenge, it was a learning experience for both the Alexa Prize team and the participants. There was no available data on how the users might use or interact with the TaskBots, and the teams had to revise their expectations. For example, many users did not invoke the TaskBots with a specific need in mind, but rather to explore this new Alexa capability, which required the teams to invest significant efforts into supporting exploration and discovery. However, the teams adapted quickly and designed engaging and innovative experiences.

% Designing for voice and screen
One of the challenges teams faced was designing experiences that harmonized voice, visuals, and touch. Our philosophy was to make TaskBot a voice-first experience, with the screen being used to supplement the voice prompts by displaying useful information that would otherwise be tricky to communicate by voice only. We connected TaskBot teams with senior Alexa UX designers for one on one consultations to learn multimodal best practices and evaluate their bot's design. Teams learned that their UX should help the user move forward with or without interacting with the screen, by providing support for key touch interactions that users expect without treating the screened device as a tablet. Another best practice was to avoid reading off everything on the screen, instead using voice to synthesize the key points and make them digestible. The most common suggestion from the UX designers was to make use of \textit{hints}, small text bubbles on the screen that provide phrases to clue the user what they can try next. Building on the learnings from the first year of the Challenge we expect many additional advances on creating natural, helpful and coherent multimodal dialog experiences.

% Designing for multiple steps and decisions
We also gained insight into how to design user experiences involving multiple steps and decisions. Users appreciated knowing up front how many total steps are in a task, and wanted to see their progress throughout the task indicating how many steps are left. To aid in decision making, sharing additional information about the different options such as user ratings and estimated duration proved helpful. 

% Conclusion
%\todo[inline]{Eugene: would be nice to include more high-level conclusions/inspirational text here}
While at the time of this writing the Finals of the challenge have not yet concluded, it is clear that the first year of the TaskBot Challenge generated many creative ideas and innovations in conversational task assistance. The Alexa Prize team and the participating university teams created a strong foundation for expanding the TaskBot Challenge and enriching the Alexa users' experience in the coming years. 

\subsubsection*{Acknowledgments}

We would like to thank all the university students and their advisors (Alexa Prize TaskBot Teams) who participated in the competition. We thank Amazon leadership and Alexa principals within the Alexa Natural Understanding (NU) and Shopping organizations for their vision and support through this entire program; Marketing (especially Shopping Marketing) for helping drive the right messaging and traffic to the Alexa Prize skill, ensuring that the participating teams received real world feedback for their research; and Alexa Engineering for the work on enabling the Alexa Prize skill. We are thankful to the Alexa Presentation Language team for providing guidance to participating teams on using APL templates to build a multimodal user experience. We are grateful to the the Alexa Developer Experience and Customer Trust (ADECT) Gadgets team for the many certification requests they worked on quickly to certify the university bots. We’d also like to thank the NU-Customer Experience team for exemplifying customer obsession by providing teams with critical inputs on building the best user experiences. We thank our leaders who took the time to virtually visit the university teams, learning from the teams and probing them to help them improve their designs. The competition would not have been possible without the support of all Alexa organizations including Speech, NLU, Data Services, Shopping Science, Conversation Modeling, and ASK and their leadership. And finally, we would like to thank Alexa users who engaged in many interactions experiencing the new Alexa Prize TaskBot and providing feedback that helped teams improve over the course of the year.

%\bibliography{custom}

%\bibliographystyle{plain}

\end{document}